%% file: main.tex
\documentclass[10pt,twocolumn,letterpaper]{article}

\usepackage{cvpr}              


\usepackage{amsmath}
\usepackage{amsfonts}
\usepackage{amssymb}
\usepackage{algorithm}
\usepackage{listings}
\usepackage{booktabs}   
\usepackage[table]{xcolor} 
\usepackage{array}      
\usepackage{booktabs}   
\usepackage{arydshln}   

\usepackage{times}
\usepackage{graphicx}
\usepackage{overpic}
\usepackage{amsmath}
\usepackage{amssymb}

\usepackage{etoolbox}
\makeatletter
\AfterEndEnvironment{algorithm}{\let\@algcomment\relax}
\AtEndEnvironment{algorithm}{\kern2pt\hrule\relax\vskip3pt\@algcomment}
\let\@algcomment\relax
\newcommand\algcomment[1]{\def\@algcomment{\footnotesize#1}}
\renewcommand\fs@ruled{\def\@fs@cfont{\bfseries}\let\@fs@capt\floatc@ruled
  \def\@fs@pre{\hrule height.8pt depth0pt \kern2pt}%
  \def\@fs@post{}%
  \def\@fs@mid{\kern2pt\hrule\kern2pt}%
  \let\@fs@iftopcapt\iftrue}
\makeatother

\newcommand{\tablestyle}[2]{\setlength{\tabcolsep}{#1}\renewcommand{\arraystretch}{#2}\centering\small}

\newcommand{\figref}[1]{Fig.~\ref{#1}}
\newcommand{\tabref}[1]{Tab.~\ref{#1}}

\usepackage{makecell}

\definecolor{cvprblue}{rgb}{0.21,0.49,0.74}
\definecolor{bbchan}{RGB}{255,130,0}
\usepackage[pagebackref,breaklinks,colorlinks,citecolor=cvprblue]{hyperref}
\definecolor{lightgray}{gray}{0.95}
\definecolor{lightpurple}{rgb}{0.9,0.9,1.0}

\usepackage{overpic}
\usepackage{booktabs}
\usepackage{multirow}
\usepackage{pifont}
\usepackage{enumitem}
\usepackage{bm}
\usepackage{diagbox}

\graphicspath{{./figs/}}
\DeclareGraphicsExtensions{.pdf,.jpg,.png}

\newcommand{\myPara}[1]{\vspace{5pt}\noindent\textbf{#1}}

\def\paperID{4368} 
\def\confName{CVPR}
\def\confYear{2026}
\definecolor{lightgreen}{RGB}{64,128,128}

\title{See What I Mean: Aligning Vision and Language Representations \\ for Video Fine-grained Object Understanding}

\author{Boyuan Sun$^{1,2}$ \quad Bowen Yin$^{1,2}$ \quad Yuanming Li$^2$ \quad Xihan Wei$^{2}$ \quad Qibin Hou$^{1,3}$\thanks{Corresponding author.} \\ 
$^1$VCIP, CS, Nankai University \qquad  $^2$Tongyi Lab, Alibaba Group  \qquad $^3$NKIARI, Shenzhen Futian \\
}

\begin{document}

\maketitle

\begin{abstract}
We present SWIM (See What I Mean), a novel training strategy that aligns vision and language representations to enable fine-grained object understanding solely from textual prompts. Unlike existing approaches that require explicit visual prompts, such as masks or points, SWIM leverages mask supervision only during training to guide cross-modal attention, allowing the model to automatically attend to the user-specified object at inference.
Our cross-attention analysis of pretrained multimodal large language models (MLLMs) reveals a systematic discrepancy: Attribute words produce sharp, localized activations in the visual modality, whereas object nouns yield diffuse and scattered patterns due to semantic reference bias and distributed high-level representations. To address this misalignment, we construct NL-Refer, an enriched dataset, in which each object mask is paired with a precise natural language referring expression. SWIM extracts multi-layer cross-attention maps from object nouns and enforces spatial consistency with ground-truth masks.  Experimental results demonstrate that SWIM substantially improves text–visual alignment and achieves superior performance over visual-prompt-based methods on fine-grained object understanding benchmarks. The code and data are available at \href{https://github.com/HumanMLLM/SWIM}{https://github.com/HumanMLLM/SWIM}.

\end{abstract}

\section{Introduction}
\label{sec:intro}
With the rapid development of large language models~\cite{qwen2, yang2025qwen3, chatgpt}, multimodal large language models (MLLMs)~\cite{Qwen-VL, xu2025qwen3, zhao2025humanomni, Qwen2-VL, wang2025internvl3_5} that can jointly reason over visual and textual modalities have recently achieved remarkable progress. 
Benefiting from large-scale pretraining on massive multimodal datasets~\cite{zhang2024video, liu2024oryx, li2016tgif, jiang2025referring}, general-purpose MLLMs~\cite{zhu2025internvl3, li2025tempsamp, vidllmsurvey} have demonstrated outstanding performance in holistic scene understanding. However, despite these impressive capabilities, they often struggle to consistently focus on user-specific objects, limiting their fine-grained object understanding abilities.

\begin{figure}[t] 
  \centering
  \setlength{\abovecaptionskip}{2pt}
  \includegraphics[width=0.98\linewidth]{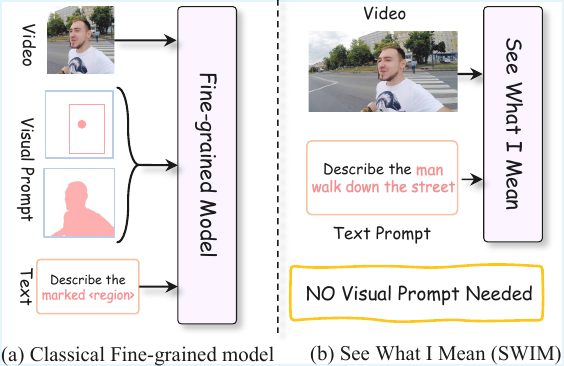}

  \caption{Pattern comparison between classical fine-grained model and SWIM. Our SWIM extracts object from pure natural language, requiring no visual prompt or any extra modules.
  } 
  \vspace{-8pt}
  \label{fig:pattern} 
\end{figure}

To enhance fine-grained object perception and understanding, a typical paradigm~\cite{zhang2024gpt4roi, zhang2025pixel, guo2024regiongpt, chen2023shikra} is to introduce additional region-level encoders that produce object-level embeddings, thereby explicitly modeling individual object tokens. In the video field, several approaches~\cite{yuan2025videorefer, peng2024inst, lin2025perceive} extend this idea by incorporating explicit visual prompts, such as points~\cite{peng2024inst}, masks~\cite{yuan2025pixelrefer}, or bounding boxes~\cite{you2023ferret}, to guide the model toward specific object regions, as shown in \figref{fig:pattern}(a).
While these approaches can successfully identify target objects through explicit visual cues, their complex designs depend on extra visual inputs, increase complexity, and diverge from the way users most naturally interact with MLLMs. In fact, specifying objects through pure natural language~\cite{chatgpt, xu2025qwen3, Qwen2.5-VL} is both more intuitive and far more common in real-world scenarios.

\begin{figure*}[t] 
  \centering
  \setlength{\abovecaptionskip}{2pt}
  \includegraphics[width=0.98\linewidth]{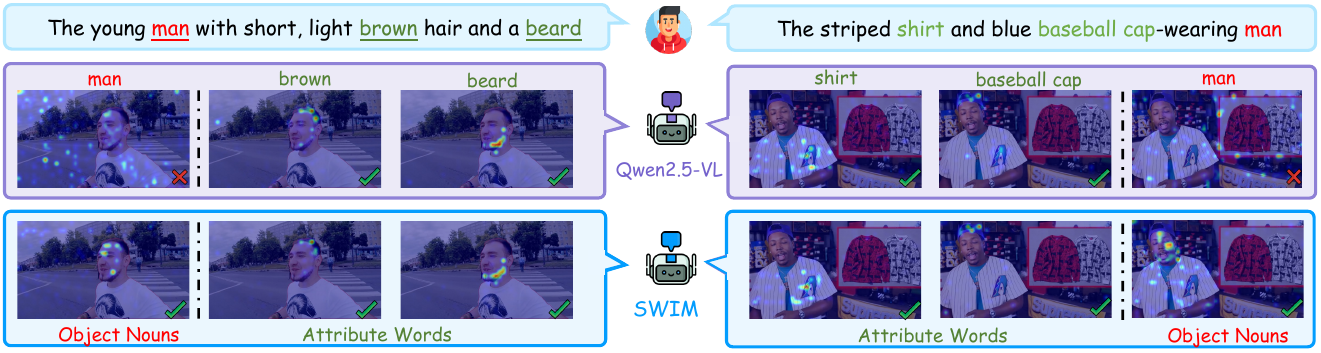}

  \caption{ Visual comparisons of cross-attention maps for object nouns and attribute words between Qwen2.5-VL~\cite{Qwen2.5-VL} and SWIM. The attribute words tied with low-level texture exhibit sharp and localized attention over correct regions, whereas the object nouns corresponding to high-level semantic concepts result in diffuse activations. This discrepancy motivates our explicit supervision strategy in SWIM.
  } 
  \vspace{-10pt}
  \label{fig:attn} 
\end{figure*}

Our motivation stems from this mismatch. As demonstrate in \figref{fig:pattern}(b), we aim to design a model that can directly locate and attend to the correct object purely from the pure textual prompt to achieve natural and fine-grained cross-modal understanding without any extra visual input in inference. 
To achieve this, we first explore how existing models attend to objects mentioned in text prompt~\cite{yuksekgonul2022and, tong2024eyes}.
Considering cross-attention between textual and visual tokens is a direct indicator of multimodal interaction~\cite{radford2021clip, yao2021filip, ding2025cross}, it can reveal whether a text token successfully grounds in a relevant visual region~\cite{zeng2025glimpse, yang2022lavt}. Therefore, by visualizing cross-attention maps for object-related words 
of Qwen2.5-VL~\cite{Qwen2.5-VL} 
in \figref{fig:attn}, we aim to uncover alignment patterns and weaknesses not apparent from standard accuracy metrics.

Interestingly, our cross-attention analysis reveals a systematic discrepancy: Attribute words~\cite{rassin2023linguistic, li2024object, ilinykh2022attention} produce sharper and more localized activations in the visual modality, while object nouns~\cite{qi2025beyond} result in diffuse and scattered attention patterns. We attribute this discrepancy to biases in semantic reference. In large-scale multimodal corpora, attribute words, such as colors or textures, correspond to specific and spatially localized visual patterns, while object nouns occur in diverse contexts, diluting their spatial association. Moreover, attribute words naturally map to low-level visual features, while object nouns rely on high-level semantic representations often varied across instances, which leads to poor alignment without explicit supervision.

This finding suggests that improving fine-grained object understanding requires explicitly strengthening cross-modal correspondence for object nouns, and thus inspires us to pursue direct supervision between object words and their associated visual regions. To provide such supervision signals, an enriched video understanding dataset that pairs object-level visual annotations with natural language prompts containing clear references is required. Thanks to earlier visual-prompt-based approaches~\cite{yuan2025videorefer, peng2024inst}, collecting training data with mask annotations is not difficult. We start from VideoRefer~\cite{yuan2025videorefer}, a video fine-grained object understanding dataset providing frame-level object masks aligned with textual prompts via placeholder tokens used for visual prompting. While these masks are valuable, the associated text does not contain clear natural language references to the objects. We therefore design a GPT-4o-powered~\cite{openai2024gpt4o} data refinement pipeline and construct the \textbf{NL-Refer} dataset. Specifically, for each placeholder, we automatically replace it with a concise natural language description of the specific 
instance, informed by the context.

Based on the NL-Refer dataset, we propose \textbf{SWIM} (\textit{See What I Mean}), a simple yet effective training strategy that explicitly aligns vision and language representations to strengthen fine-grained object understanding in MLLMs. Specifically, during supervised fine-tuning, SWIM extracts cross-attention maps for object nouns from multiple intermediate layers and aligns them with ground-truth object masks, enforcing spatial consistency between textual identity and visual grounding. By providing explicit alignment signal throughout training, SWIM guides the model to preserve and utilize fine-grained object-level information, enabling more precise visual localization from purely textual prompts at inference. Extensive experiments across fine-grained object understanding benchmarks demonstrate that SWIM enhances text–visual 
alignment and outperforms visual-prompt-dependent approaches.

We summarize our contributions as follows.
\begin{itemize}
\item We point out design limitations in existing fine-grained object understanding models and the insufficient vision–language alignment in general MLLMs. Based on the observed systematic discrepancy, we introduce NL-Refer dataset, in which each object is referred with explicit natural language expressions.
\item We propose a novel training strategy, SWIM (See What I Mean), which explicitly enforces alignment between visual content and object nouns during training, resulting in a model that requires no visual prompt inputs at inference.
\item Through experiments on fine-grained object understanding benchmarks, SWIM demonstrates consistent improvements over visual-prompt-based approaches. Quantitative and qualitative analyses of text–visual alignment further corroborate our claim.
\end{itemize}

\begin{figure*}[t] 
  \centering
  \setlength{\abovecaptionskip}{2pt}
  \includegraphics[width=0.98\linewidth]{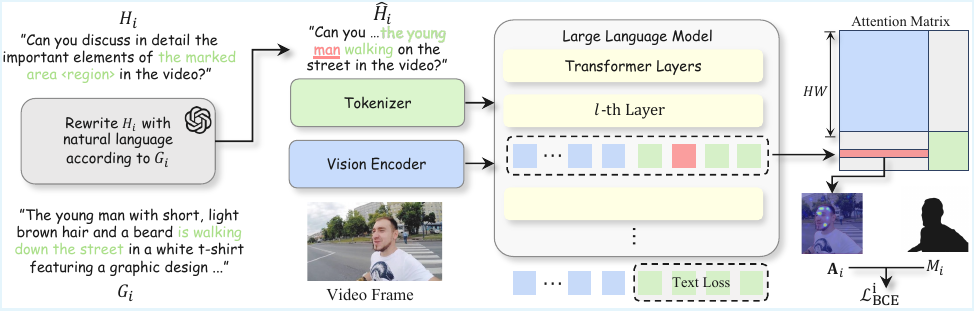}

  \caption{ \textbf{Training pipeline of SWIM.} Explicitly supervision is applied on cross-attention between object noun and visual tokens, enable accurate fine-grained object grounding from pure neutral text prompts at inference without any extra visual prompt.
  } 
  \label{fig:method} 
\end{figure*}

\section{Related Work}
\label{sec:relatedwork}
\subsection{Multimodal Large Language Model}
Multimodal large language models (MLLMs)~\cite{yang2025humanomniv2, liu2023llava, comanici2025gemini, chen2024sharegpt4video, fu2024mme} integrate visual signals with textual inputs, leveraging the powerful reasoning and generative capabilities of LLMs~\cite{liu2024deepseek, team2025kimi, dubey2024llama, guo2025deepseek, minaee2024large, chen2024far} to tackle a wide range of tasks~\cite{zhao2025facial, wang2025x, jin2026geoagent}. Beyond image-based approaches~\cite{ lin2024vila,liu2023improved}, recent advances in spatiotemporal architectures design~\cite{videollamb, damonlpsg2023videollama, damonlpsg2025videollama3}  enables MLLMs to extend multimodal understanding into the video filed~\cite{zhang2024long, sun2025llava,ning2025video}, achieving strong performance in real-world applications~\cite{duan2025docopilot, hu2025flowsearch}. 

However, despite advances, MLLMs still face challenges in fine-grained object understanding, especially when identifying or describing user-specified targets from solely textual prompts~\cite{aligning_llm_human}.
One potential reason lies in the issue of vision–language alignment within such models~\cite{wang2024reconstructive, zhang2023tale}. 
As shown in Fig.~\ref{fig:attn}, the discrepancy that attribute words tend to produce clear attention patterns, while object nouns often result in diffuse and scattered activations motivates us to design SWIM, which applies supervision on cross-modal correspondence of specific textual tokens. 
Although prior studies~\cite{kaduri2025s, kang2025your, zhang2025cross, jiang2025devils, sun2025depth} examined intermediate feature representations in MLLMs, and works like Cambrain~\cite{tong2024cambrian} and VIRAL~\cite{yoon2025visual} explores reconstructing visual features from intermediate layers, they generally focus on visual embeddings and ignore the visual-language alignment.
\subsection{Fine-grained Object Understanding}
To enhance the fine-grained understanding capability of MLLMs, recent approaches~\cite{setofmark,  zhang2024omg, chen2023position, huang2024segment, zhang2024ferret, xuan2024pink, yue2024sc, zhao2023chatspot, rasheed2024glamm, cai2023making, zhan2024griffon, fei2024vitron, wu2025spatial} start to focus on object-centric perception and reasoning. 
The most common paradigm~\cite{zhang2024gpt4roi, yuan2024osprey, guo2024regiongpt, you2023ferret, tian2024chatterbox,yu2025merlin, OmniRGPT} is to involve explicit visual prompts (points/boxes/masks) as guidance and additional encoders to improve fine-grained comprehension of local regions. 
For example, VideoRefer~\cite{yuan2025videorefer} and PixelRefer~\cite{zhang2025pixel} employ masks as visual indicators and leverage extra visual encoders to enhance the perception of objects. They also contribute valuable video fine-grained datasets with mask annotations. 

In a word, these methods require additional encoders and extra visual prompt~\cite{yan2024list, zhao2024mg, cai2024vip} even at inference time, which brings extra computational cost and deviates from typical user interaction patterns. 
Meanwhile, SWIM adopts a training paradigm with explicit supervision for cross-modal alignment, enabling fine-grained object understanding from natural language references without architecture modifications or any extra visual prompts at inference.


\section{See What I Mean (SWIM)}
\subsection{NL-Refer: Dataset Construction}

To provide explicit supervision for aligning object nouns in text with their corresponding visual regions, we construct \textbf{NL‑Refer} from the VideoRefer dataset, which can be represented as
\begin{equation}
\mathcal{D}_{\mathrm{VideoRefer}} = \big\{\, (V_i, H_i,\, G_i,\, M_i) \;\big|\; i = 1,\dots,N \,\big\},
\end{equation}
where $V_i$ denotes the video, $H_i$ denotes the “human” message containing the placeholder token $ \texttt{<region>}$ , $G_i$ denotes the paired “gpt” response describing the marked object in natural language, and $M_i$ denotes the corresponding pixel-level instance masks for the target region.

While such placeholders support visual-prompt-based methods, they do not convey the explicit semantic identity of the referenced object, thereby inhibiting a model’s ability to learn robust and direct text–visual correspondences for object nouns. 
To mitigate this, we propose a refinement process leverages GPT‑4o to replace each placeholder $ \texttt{<region>}$ in the human message $H_i$ with a concise and unambiguous natural language referring expression $r_i$ that specifies the object instance, using descriptive details drawn from the paired ``gpt'' response $G_i$. This generation process maintains the original conversational structure while embedding explicit semantic content in the prompt. Within each $r_i$, GPT‑4o identifies the single most representative object noun $w_i$, the lexical item that captures the core semantic identity of the target, and surrounds it with a special markup token \texttt{<ins>} to enable deterministic location of the corresponding token for further training. This inline tagging is directly linked to $M_i$, so that the marked noun token is aligned with the ground-truth mask it denotes.

Formally, let $H_i$ denote the original human message and $G_i$ the corresponding model response in VideoRefer, paired with mask $M_i$. The refined human message $\hat{H}_i$ is given by
\begin{equation}
\hat{H}_i = \mathrm{Mark}\!\left(
\mathrm{Replace}\!\left(H_i,\, \texttt{<region>},\, r_i\right),\,
w_i
\right),
\end{equation}
where $\mathrm{Replace}(\cdot)$ substitutes the placeholder with the GPT‑4o-generated referring expression $r_i$, and $\mathrm{Mark}(\cdot, w_i)$ encloses $w_i$ in \texttt{<ins>} delimiters. The referring expression itself is obtained as
\begin{equation}
r_i = \mathrm{NLRef}(G_i),
\end{equation}
where $\mathrm{NLRef}(\cdot)$ extracts salient object descriptors from $G_i$ and composes them into a minimal, discriminative phrase. 

The refined dataset is then defined as
\begin{equation}
\mathcal{D}_{\mathrm{NL\!-\!Refer}} = \big\{\,(V_i, \hat{H}_i,\, G_i,\, M_i) \;\big|\; i = 1,\dots,N \,\big\},
\end{equation}
with $\hat{H}_i$ containing the linguistically explicit object reference, $G_i$ providing the unchanged descriptive context, and $M_i$ denoting the ground-truth mask of the target instance. By systematically embedding object identity into the text and marking its precise token span, $\mathcal{D}_{\mathrm{NL\!-\!Refer}}$ establishes a reliable mapping between lexical items and visual annotations, laying a solid foundation for subsequent cross-attention supervision of object nouns.

\begin{table*}[ht]
\centering
\setlength{\abovecaptionskip}{2pt}
\small
\tablestyle{6.3pt}{1.0}
\caption{Performance comparisons on fine-grained VideoRefer-Bench-D and VideoRefer-Bench-Q. 
The best results are \textbf{bold} and the second-best results are \underline{underlined}.}
\begin{tabular}{lcccccccccccc}
\toprule
 &   & \multicolumn{6}{c}{\textbf{VideoRefer-Bench-Q}} & \multicolumn{5}{c}{\textbf{VideoRefer-Bench-D}} \\
\cmidrule(lr){3-8} \cmidrule(lr){9-13}
\textbf{Method} & Prompt types & \textbf{Basic} & \textbf{Seq.} & \textbf{Rel.} & \textbf{Rea.} & \textbf{Fut.} & \textbf{Avg.} 
 & \textbf{SC} & \textbf{AD} & \textbf{TD} & \textbf{HD} & \textbf{Avg.} \\
\hline
\rowcolor{lightgray}
\multicolumn{12}{c}{\textit{Generalist Models}} \\
\hline

LongVU-7B~\cite{shen2024longvu} & Text  & 47.2 & 61.3 & 57.5 & 85.3 & 65.8 & 61.0 & 2.33 & 1.80 & 2.39 & 1.68 & 2.05 \\
LongVA-7B~\cite{zhang2024long} &  Text  & 56.2 & 62.5 & 52.0 & 83.9 & 65.8 & 61.8 & 3.02 & 2.30 & 1.92 & 2.51 & 2.44 \\
LLaVA-OV-7B~\cite{li2024llavaone} & Text & 58.7 & 62.9 & \underline{64.7} & 87.4 & 76.3 & 67.4 & 3.09 & 1.94 & 2.50 & 2.41 & 2.48 \\
Qwen2-VL-7B~\cite{Qwen2-VL} & Text  & 62.0 & 69.6 & 54.9 & 87.3 & 74.6 & 66.0 & 3.99 & 3.05 & 2.44 & 2.44 & 2.97 \\
Qwen2.5-Omni-7B~\cite{Qwen2.5-Omni} & Text  & 65.0 & 67.6 & 54.0 & 84.3 & 70.3 & 68.2 & 3.92 & 2.82 & 1.97 & 2.34 & 2.76 \\
InternVL2-26B~\cite{chen2023internvl}&  Text & 58.5 & 63.5 & 53.4 & 88.0 & \underline{78.9} & 65.0 & 4.08 & 3.35 & 3.08 & 2.28 & 3.20 \\
Qwen2.5-VL-7B~\cite{Qwen2.5-VL}& Text & \underline{78.0} & \underline{69.7} & 58.2 & 79.9 & 73.2 & 71.8 & 4.33 & 3.19	& 2.88 & 2.58 & 3.24\\
GPT-4o-mini~\cite{openai2024gpt4o} & Text  & 57.6 & 67.1 & 56.5 & 85.9 & 75.4 & 65.8 & 3.89 & 3.18 & 2.62 & 2.50 & 3.05 \\
GPT-4o~\cite{openai2024gpt4o}  &   Text  & 62.3 & 74.5 & 66.0 & 88.0 & 73.7 & 71.3 & 4.15 & 3.31 & 3.11 & 2.43 & 3.25 \\
\hline
\rowcolor{lightgray}
\multicolumn{12}{c}{\textit{Specialist Models}} \\
\hline
Elysium-7B~\cite{wang2024elysium} & Box   & -- & -- & -- & -- & -- & -- & 2.35 & 0.30 & 0.02 & \textbf{3.59} & 1.57 \\
Artemis-7B~\cite{qiu2024artemis}& Box & -- & -- & -- & -- & --  & -- & 3.42 & 1.34 & 1.39 & 2.90 & 2.26 \\
Osprey-7B~\cite{yuan2024osprey}& Point, Mask & 45.9 & 47.1 & 30.0 & 48.6 & 23.7 & 39.9 & 3.30 & 2.66 & 2.10 & 1.58 & 2.41 \\
Ferret-7B~\cite{you2023ferret} &  Point, Box, Mask  & 35.2 & 44.7 & 41.9 & 70.4  & 74.6 & 48.8 & 3.20 & 2.38 & 1.97 & 1.38 & 2.23 \\
PAM-3B~\cite{lin2025perceive}& Point, Box, Mask & -- & -- & -- & -- & -- & -- & 3.92 & 2.84 & 2.88 & 2.94 & 3.14\\

DAM-8B~\cite{lian2025describe} & Point, Box, Mask & -- & -- & -- & -- & --  & -- &  \underline{4.69} & \underline{3.61} & \underline{3.34} & 3.03 & \underline{3.68} \\
VideoRefer-7B~\cite{yuan2025videorefer} & Mask & 75.4 & 68.6 & 59.3 & \underline{89.4} & 78.1  & \underline{71.9} & 4.44 & 3.27 & 3.10 & \underline{3.04} & 3.46 \\
 \rowcolor[HTML]{fff5f4}
 \textbf{SWIM}& Text & \textbf{83.8} & \textbf{75.0} & \textbf{66.7} & \textbf{93.7} & \textbf{80.7} & \textbf{78.3} & \textbf{4.92} & \textbf{3.85} & \textbf{3.43} & 2.96 & \textbf{3.78}\\
\bottomrule
\end{tabular}
\label{tab:videorefer}
\end{table*}

\begin{table}[t]
    \centering
    \tablestyle{2.5pt}{1.0}
    \setlength{\abovecaptionskip}{2pt}
    \small
    \caption{
    Performance comparisons on general video benchmarks.
    }
    \begin{tabular}{lcccccc} \toprule
    \textbf{Method} & \textbf{MVBench} & \textbf{VideoMME} & \textbf{ActivityNet} \\
    \midrule
    VideoLLaMA2~\cite{damonlpsg2024videollama2} & 54.6 & 47.9 & 50.2\\
    VideoChat2-HD~\cite{li2023videochat}& 51.1 & 54.6 & 49.1\\
    VideoLLaMA2.1~\cite{damonlpsg2024videollama2} & 57.3 & 54.9 & -\\
    LLaVA-Next-Video~\cite{zhang2024llavanext-video} & - & 46.5 & 53.5\\
    LLaVA-Octopus~\cite{zhao2025llava} & 51.7 & 55.7 & 53.4\\
    INST-IT~\cite{peng2024inst} & - & 54.0 & 55.2\\
    VideoRefer~\cite{yuan2025videorefer} & 59.6 & 55.9 & -\\
    \textbf{SWIM} & \textbf{62.1} & \textbf{55.9} & \textbf{55.6}\\
    \bottomrule
    \end{tabular}
    \label{general benchmark}
\end{table}

\subsection{Attention Regularization}

Leveraging $\mathcal{D}_{\mathrm{NL\!-\!Refer}}$, in which each refined textual prompt $\hat{H}_i$ contains precisely one object noun $w_i$ tagged with \texttt{<ins>} delimiters and deterministically linked to the ground-truth mask $M_i$, we design an auxiliary supervision mechanism to explicitly align the visual grounding of noun tokens with their annotated object regions. As analyzed in Section~\ref{sec:intro}, the cross-attention patterns in existing MLLMs exhibit an empirical vision–language misalignment: Object nouns often produce diffuse and scattered activations across the visual tokens. This systematic discrepancy motivates us to directly guide the model’s cross-modal attention when processing tagged object noun tokens, so that their activation is concentrated on the relevant visual region.

During training, we first tokenize $\hat{H}_i$ into a sequence of $L_t$ tokens, yielding text embeddings $\mathbf{X}^t \in \mathbb{R}^{L_t \times d}$, where $d$ denotes the hidden dimension. Let $j_i \in \{1,\dots,L_t\}$ be the index corresponding to the tagged noun token $w_i$. Within the LLM decoder, these text embeddings interact with a sequence of $L_v$ visual tokens $\mathbf{X}^v \in \mathbb{R}^{L_v \times d}$ through cross-attention layers.
For a given cross-attention module at layer index $l$, let $\mathbf{Q}^t_{l}[j_i] \in \mathbb{R}^{d}$ be the query vector of the tagged noun token at position $j_i$, and let $\mathbf{K}^v_{l} \in \mathbb{R}^{L_v \times d}$ denote the key vectors of the all visual tokens of one frame from $V_i$. The cross-attention weights from $w_i$ to the visual tokens at layer $l$ are computed as:
\begin{equation}
\mathbf{A}_{l,i} = \mathrm{softmax}\!\left( \frac{\mathbf{Q}^t_l[j_i]\, (\mathbf{K}^v_l)^\top}{\sqrt{d}} \right) ,
\end{equation}
where the softmax is applied over the $L_v$ visual token positions. Each element indicates the degree to which the noun token attends to each visual token at layer index $l$.

To enable spatial supervision, the attention vector $\mathbf{A}_{l,i}$ is mapped to the original feature grid of resolution $(H, W)$ that aligns with $M_i$. 
This mapping follows the spatial correspondence between visual tokens and encoder patches.
If $(H, W)$ differs from the token grid resolution, bilinear interpolation is applied to match the mask resolution exactly. 
The resulting attention map for layer $l$ is denoted as $\mathbf{A}_{l,i} \in [0,1]^{H \times W}$. Since attention patterns may vary across layers, we aggregate attention maps from a selected set of layers $\mathcal{S}$ by simple averaging:
\begin{equation}
\bar{\mathbf{A}}_i = \frac{1}{|\mathcal{S}|} \sum_{l \in \mathcal{S}} \mathbf{A}_{l,i}.
\end{equation}
The aggregated map $\bar{\mathbf{A}}_i$ captures the stable cross-modal correspondence between the tagged object noun $w_i$ and its visual region after accounting for multi-layer variability.

Finally, we supervise $\bar{\mathbf{A}}_i$ with the binary mask $M_i$ using a pixel-wise binary cross-entropy loss:
\begin{equation}
\begin{aligned}
\mathcal{L}_{\mathrm{BCE}}^{(i)} = 
 & -\frac{1}{HW} \sum_{u=1}^H \sum_{v=1}^W \Big[ M_i(u,v) \,\log \bar{\mathbf{A}}_i(u,v) \\
 & \quad + \big(1 - M_i(u,v)\big) \,\log\big( 1 - \bar{\mathbf{A}}_i(u,v) \big) \Big],
\end{aligned}
\end{equation}
where $M_i(u,v) \in \{0,1\}$ indicates whether pixel $(u,v)$ belongs to the target object. By providing this explicit alignment signal at training time, the model learns to consistently concentrate cross-attention from object nouns onto their correct visual regions, bridging the alignment gap identified in our analysis and enhancing fine-grained understanding performance without modifying the base architecture. 

Notably, unlike many existing fine-grained object understanding approaches that require the visual prompt mask $M_i$ as part of the inference input, in our SWIM, $M_i$ is only used for attention regularization during supervised fine-tuning, incurring no additional burden in inference.

\begin{table*}[t]
    \centering
    \small
    \setlength{\abovecaptionskip}{2pt}
    \definecolor{lightlightgray}{gray}{0.8}
    \tablestyle{4pt}{1.0}
    \caption{Ablation study of attention layer selection among different layer number and layer index.}
    \begin{tabular}{cccccccc}
        \toprule
        \textbf{Layer Number} & \textbf{Layer Index} &  \textbf{VideoRefer-D} & \textbf{Layer Number}  & \textbf{Layer Index} & \textbf{VideoRefer-D} \\
        \midrule
        1 & [1] & 3.43 & 6 & [1, 3, 5, 7, 9, 11] & 3.73 \\
        1 & [13] & 3.48  & 6 &  [1, 6, 11, 16, 21, 26] & 3.78 \\
        1 &[27]& 3.52  & \cellcolor[HTML]{fff5f4}6 & \cellcolor[HTML]{fff5f4}[2, 7, 12, 17, 22, 27] & \cellcolor[HTML]{fff5f4}3.78 \\
        3 & [1, 13, 27] & 3.72 & 6 & [17, 19, 21, 23, 25, 27] & 3.76 \\
        3 & [9, 18, 27] & 3.70 & 9 & [1, 4, 7, 10, 13, 16, 19, 22, 25] & 3.75\\
        3 & [1, 7, 13] & 3.69&  14 & [1, 3, 5, 7, 9, 11, 13, 15, 17, 19, 21, 23, 25, 27] & 3.77 \\
        
        \bottomrule
    \end{tabular}
    \vspace{-5pt}
    \label{tab: layer}
\end{table*}

\begin{table}[t]
    \centering
      \tablestyle{10.9pt}{1.0}
      \setlength{\abovecaptionskip}{2pt}
      \small
    \caption{
    Ablation study of attention layer fusion methods on VideoRefer-Bench-D. Prod. denotes element-wise product.
    }
    \begin{tabular}{lcccccc} \toprule
\textbf{Fusion} & \makecell{\textbf{Subject} \\ \textbf{Correspondence} } 
& \makecell{\textbf{Temporal } \\ \textbf{Description}}  
& \textbf{Avg.}\\
    \midrule
    Add & 4.62 & 3.24 & 3.57 \\
    Pool & 4.56 & 3.11 & 3.49 \\
    Prod. & 4.81 & 3.21 & 3.55 \\
    \textbf{Mean} & \textbf{4.92} & \textbf{3.43} & \textbf{3.78}\\
    \bottomrule
    \end{tabular}
    \label{sample method}
\end{table}

\section{Experiments}

\subsection{Experimental Settings}
\myPara{Implementation details.} We implement SWIM on top of the widely used Qwen2.5VL-7B~\cite{Qwen2.5-VL} framework, which employs SIGLIP (so400m-patch14-384)~\cite{zhai2023sigmoid} as the visual encoder and Qwen2.5~\cite{qwen2.5} as the large language model (LLM) decoder. Our training set is composed of two parts: (1) The proposed \textbf{NL-Refer} dataset, converted from the detailed caption subset of VideoRefer-700K dataset~\cite{yuan2025videorefer}, containing 125K videos with refined textual annotations that explicitly refer to objects in natural language along with their corresponding instance masks; (2) A portion of general video-based QA data from LLaVA-Video-178K~\cite{zhang2024video} and videorefer-qa-75k. For the LLaVA-Video-178K, we decompose multi-turn dialogues into single-turn QA pairs (1.3M in total), and sample 100K QA pairs (approximately $7.5\%$) for training. We also sample 10K QA pairs from videorefer-qa-75k to maintain its ability on multi-choice question.  In total, our training data contains 235K examples, which is significantly smaller than that used for most generalist MLLMs and is less than 1/3 of the VideoRefer. 
All experiments are conducted on $8\times$~NVIDIA A100 GPUs.

\myPara{Evaluation benchmarks.}
To demonstrate the effectiveness of SWIM, we evaluate it from both fine-grained video object understanding and general video understanding perspectives. We adopt \textbf{VideoRefer-Bench}~\cite{yuan2025videorefer}, a dedicated benchmark for object-level video understanding that comprises two sub-tasks. \textbf{VideoRefer-Bench-D} measures description generation for specified objects, containing 400 curated entries from Panda-70M~\cite{chen2024panda}. Outputs are scored from 0-5 in four aspects: Subject Correspondence (SC, subject matches ground truth), Appearance Description (AD, accuracy of color/shape/texture), Temporal Description (TD, correctness of motion), and Hallucination Detection (HD, absence of invented details). \textbf{VideoRefer-Bench-Q} evaluates object-level understanding and reasoning, consisting of 198 videos from DAVIS-2017~\cite{pont20172017} and MeViS~\cite{ding2023mevis, ding2025mevis}, paired with 1{,}000 region-linked multiple-choice questions spanning Basic (simple factual queries), Sequential (temporal order reasoning), Relationship (relations between objects), Reasoning (context-based inference), and Future (predict future states).

As for general video understanding, we adopt three representative benchmarks: \textbf{MVBench}~\cite{li2024mvbench}, which offers diverse multi-aspect evaluations of video-language reasoning; \textbf{Video-MME}~\cite{fu2024videomme}, a comprehensive suite covering spatio-temporal reasoning, event localization, and attribute recognition; and \textbf{ActivityNet-QA}~\cite{yu2019activityqa}, a large-scale QA dataset based on ActivityNet videos targeting a wide range of skills from event recognition to temporal reasoning. These benchmarks jointly examine SWIM’s generalization ability beyond fine-grained object grounding.

\subsection{Main Results}

\subsubsection{Results on Fine-grained Benchmarks}
In \tabref{tab:videorefer}, we summarize the performance of SWIM and a range of generalist and specialist state-of-art approaches on the fine-grained VideoRefer-bench~\cite{yuan2025videorefer}, which evaluates video fine-grained object understanding in both question-answering and description settings. 

On \textbf{VideoRefer-Q}, which evaluates fine-grained object understanding through five sub-tasks, SWIM attains substantial gains in Basic (+5.8$\%$ over Qwen2.5-VL-7B~\cite{Qwen2.5-VL}) and Sequential (+5.3$\%$) cases, which demand precise object identification before answering. SWIM yields an average accuracy of 78.3$\%$, exceeding the VideoRefer-7B~\cite{yuan2025videorefer} by +6.4$\%$, and surpassing all generalist models such as Qwen2.5-VL-7B (71.8) and GPT-4o (71.3)~\cite{openai2024gpt4o}.

On \textbf{VideoRefer-D}, which assesses spatial correspondence (\textbf{SC}), action description (\textbf{AD}), temporal description (\textbf{TD}), and higher-level human–object description (\textbf{HD}), SWIM achieves 4.92, 3.85, 3.43, and 2.96 respectively, for an average of 3.78, outperforming the best specialist baseline DAM-8B (3.68) and the strongest generalist system GPT-4o (3.25). The performance gain on SC (+0.23) and AD (+0.24) over DAM-8B~\cite{lian2025describe} highlights the strength of SWIM in aligning object nouns to precise instance regions.

Overall, SWIM delivers consistent improvements across both QA and description tasks, indicating more precise referring capability between natural language and visual regions. By integrating explicit attention regularization alignment supervision at training only, SWIM enhances the fine-grained grounding capability of MLLMs without incurring architectural changes or inference-time visual prompting, making it competitive across diverse evaluation scenarios that demand high-resolution text–visual alignment.

\subsubsection{Results on General Benchmarks}
Other than the fine-grained understanding benchmarks, we also evaluate SWIM on several representative general video understanding benchmarks, including MVBench~\cite{li2024mvbench}, Video-MME~\cite{fu2024videomme}, and ActivityNet-QA~\cite{yu2019activityqa}, as summarized in \tabref{general benchmark}. The results indicate that while SWIM is primarily optimized for fine-grained video object understanding tasks, its performance on broader video-language understanding tasks remains within a competitive range compared to existing methods. This suggests that the proposed training strategy for cross-attention alignment does not substantially compromise general video understanding ability, allowing SWIM to retain acceptable performance in more comprehensive scenarios.

\begin{table}[t]
\centering
\tablestyle{11pt}{1.0}
\setlength{\abovecaptionskip}{2pt}
\small
\caption{Ablation study of attention loss function used in SWIM on VideoRefer-Bench-D.}
\begin{tabular}{lcccccc}
\toprule
\textbf{Loss} 
& \makecell{\textbf{Subject} \\ \textbf{Correspondence} } 
& \makecell{\textbf{Temporal } \\ \textbf{Description}}  
& \textbf{Avg.}\\
    \midrule
    mIoU & 4.88 & 3.34 & 3.71 \\
    Focal & 4.80 & 3.24 & 3.69 \\
    Dice & 4.90 & 3.38  & 3.74 \\
    \textbf{BCE} & \textbf{4.92} & \textbf{3.43} & \textbf{3.78}\\
    \bottomrule
\end{tabular}
\label{loss}
\vspace{-10pt}
\end{table}

\subsection{Ablation Analysis}
\subsubsection{Effect of Attention Layer Selection}
We first explore the influence of the number and positions of cross-attention layers used for supervision in SWIM. As shown in \tabref{tab: layer}, we evaluate VideoRefer-D performance under configurations ranging from single-layer supervision to selecting up to 14 layers. We find that increasing the number of supervised layers yields consistent improvements initially. The performance rises from $3.43$ at a single shallow layer to $3.78$ when supervising six layers. Beyond six layers, results tend to be stable with all larger configurations remaining within $0.02$ of each other. 
Furthermore, evenly spaced selection of supervision layers across the network produces better or comparable results than densely clustered layers in a narrow depth range, indicating that a balanced distribution from early to late stages fosters more stable cross-modal alignment. This observation suggests that SWIM achieves the best trade-off with moderately deep and uniformly spaced supervision layers.

\begin{figure}[t] 
  \centering
  \setlength{\abovecaptionskip}{2pt}
  \includegraphics[width=0.95\linewidth]{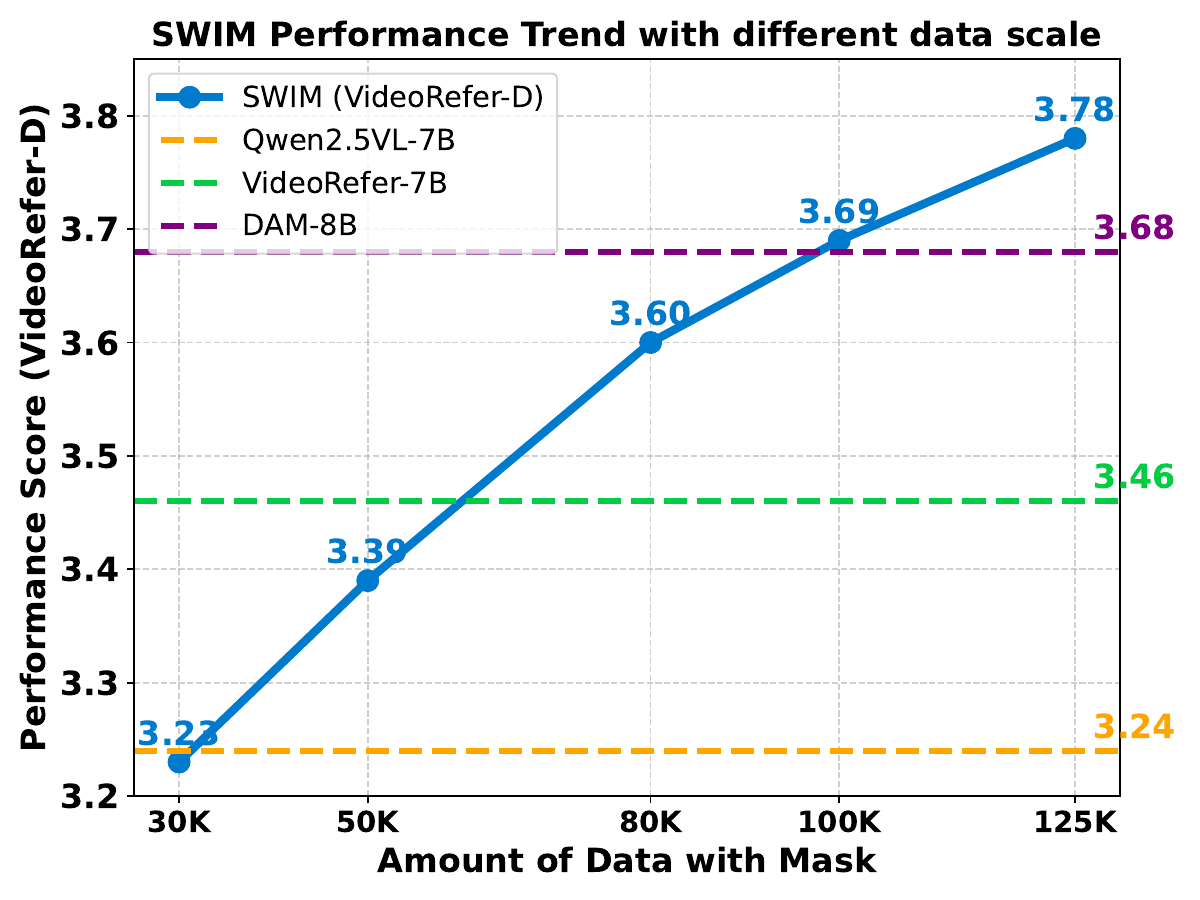}

  \caption{\textbf{Scalablity of SWIM.} The performance of SWIM scales consistently with the increase in data scale.
  } 
  \label{fig:data_scale} 
\end{figure}

\subsubsection{Effect of Attention Layer Fusion}
We further study how attention maps extracted from multiple layers should be fused to provide the alignment signal in SWIM. Several fusion strategies are considered, including addition, pooling, mean, and element-wise product. 
As shown in \tabref{sample method}, simple mean aggregation yields the highest average score (3.81), exceeding addition (3.57) and pooling (3.49) by a clear margin.  This can be attributed to its ability to preserve consistent spatial patterns across layers without introducing bias toward any single depth, effectively smoothing noise and retaining salient activation peaks. In contrast, The element-wise product requires all layers to highlight the same locations to retain them, thus tends to over-suppress regions with moderate but meaningful attention, reducing overall coverage of the target object.

\subsubsection{Effect of Loss Function}
We also examine the impact of the loss function used to supervise cross-attention in SWIM. 
As demonstrated in \tabref{loss}, compared to alternatives such as mIoU, Focal, and Dice losses, binary cross-entropy (BCE) consistently yields superior overall performance. This advantage can be attributed to the sparsity inherent in attention maps extracted from MLLMs. When aligning object nouns to visual regions, activations are highly localized and occupy only a small fraction of the spatial grid due to the softmax operation. Loss functions that emphasize overlap ratios or focus disproportionately on hard negatives may under-penalize diffuse activations, making it harder to enforce precise alignment. In contrast, BCE treats each pixel independently and applies a uniform probabilistic penalty across all spatial locations, encouraging suppression of irrelevant high-activation regions while reinforcing confident attention on the target area. This balanced penalization aligns well with the fine-grained supervision signals in SWIM, leading to more accurate and stable grounding of object nouns.

\subsection{Scalability with Mask-Annotated Data Volume}

Beyond the intrinsic performance of the model, scalability is also a critical property for multimodal models. 
In the field of fine-grained object understanding, scalability determines whether stronger alignment can be achieved simply by expanding high-quality annotated datasets, making it essential for long-term advancement.
Therefore, to examine SWIM’s scalability, we vary the number of mask-annotated training videos from NL-Refer dataset. The dataset size is gradually expanded from a 30K subset to the maximum available 125K samples. \figref{fig:data_scale} shows the overall alignment score at each scale, revealing a clear and monotonic upward trend.

The results demonstrate that SWIM can effectively leverage additional fine-grained supervision. Each increase in mask-annotated data yields measurable gains, and the improvement persists up to the largest scale tested. This sustained growth can be attributed to the explicit alignment between object nouns and masks in our training pipeline, which enables the model to refine cross-modal attention alignments across diverse examples without overfitting to narrower data distributions.

In addition, the absence of a plateau at 125K data scale indicates that SWIM is inherently capable of benefiting from larger-scale mask supervision. Although our experiments are bounded by the current data availability, the persistent upward trajectory suggests considerable untapped potential. SWIM may achieve stronger alignment if broader and more diverse mask-annotated corpora are provided.

\begin{table}[t]
    \centering
    \small
    \setlength{\abovecaptionskip}{2pt}
    \definecolor{lightlightgray}{gray}{0.8}
    \tablestyle{6.4pt}{1.0}
    \caption{\textbf{GamePoint@P} \textbf{between Qwen2.5-VL and SWIM.} This metric measures the visual region with the highest cross-attention score falls within the specified object mask.}
    \begin{tabular}{lccccccc}
        \toprule
        \textbf{Method} & \textbf{G.P.@P-1} &  \textbf{G.P.@P-5} & \textbf{G.P.@P-10}   \\
        \midrule
        Qwen2.5-VL-7B & 0.329 & 0.293 & 0.270  \\
        \textbf{SWIM} & \textbf{0.392} & \textbf{0.348} & \textbf{0.317} \\
        
        \bottomrule
    \end{tabular}
    
    \label{tab:gamepointP}
\end{table}

\begin{figure*}[t] 
  \centering
  \setlength{\abovecaptionskip}{2pt}
  \includegraphics[width=0.95\linewidth]{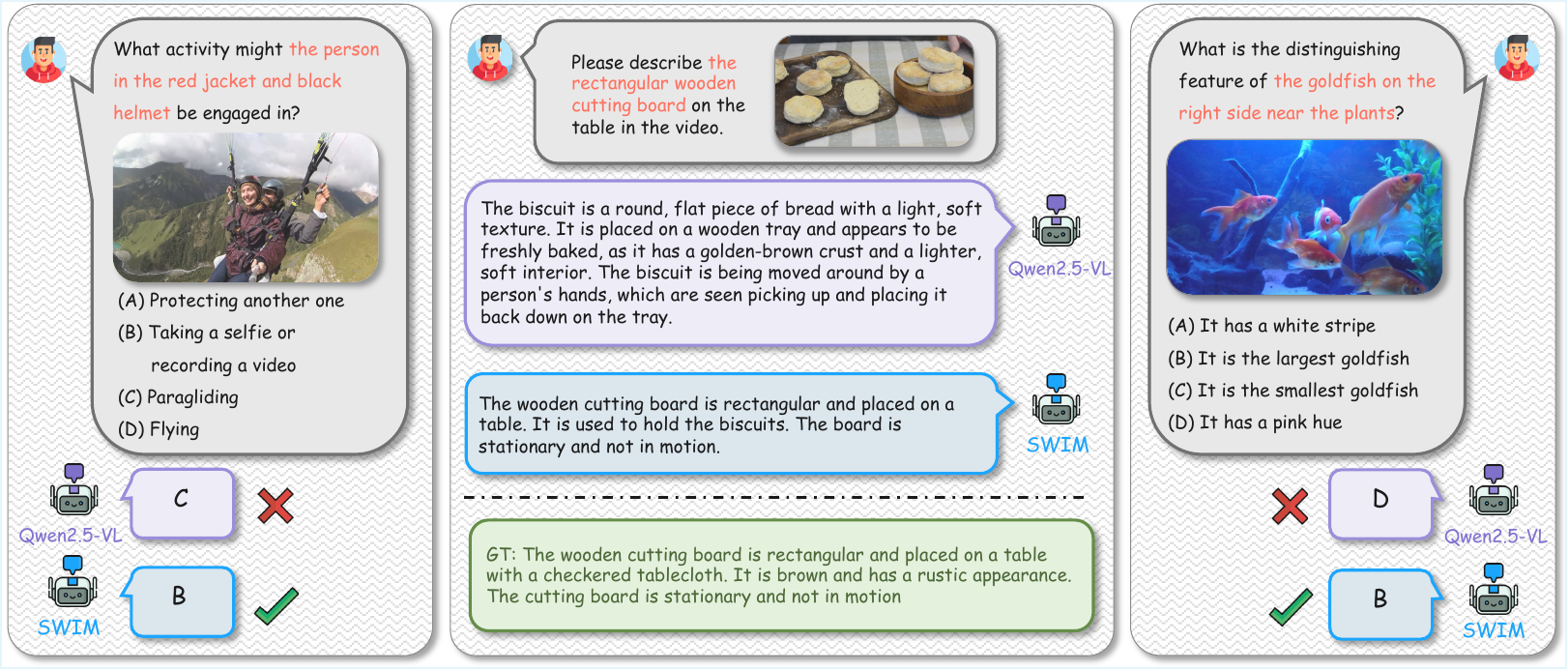}

  \caption{Qualitative comparisons between SWIM and Qwen2.5-VL~\cite{Qwen2.5-VL}. 
  } 
  \vspace{-10pt}
  \label{fig:qualitative} 
\end{figure*}

\subsection{GamePoint-based Attention Localization}

We evaluate spatial grounding using the \textbf{GamePoint@P} metrics, as reported in \tabref{tab:gamepointP}. 
GamePoint@P measures the proportion for the top $P\%$ attention pixels highest-attention pixels in $\bar{\mathbf{A}}_i$ that fall within the object mask $M_i$:
\begin{equation}
\mathrm{GamePoint@P} = \frac{1}{N} \sum_{i=1}^{N} \frac{|\mathrm{TopPerc}(\bar{\mathbf{A}}_i, P) \cap P_i|}{|\mathrm{TopPerc}(\bar{\mathbf{A}}_i,P)|},
\end{equation}
where $P_i = \{(u,v) \mid M_i(u,v) = 1\}$ denotes object-region pixels, $\mathrm{TopPerc}(\cdot)$ denotes selecting the top $P\%$ elements.

From \tabref{tab:gamepointP}, SWIM consistently outperforms Qwen2.5-VL-7B across all $P$. The improvement is most pronounced at $P=1$ (+6.3\%) and $P=5\%$ (+5.5\%), showing that SWIM’s most confident attention points and top regions are far more likely to land within the correct object area. 
This suggests SWIM achieves sharper and more focused attention peaks on the target object, raising coverage across all top-$P$ settings, and overcoming the diffuse attention patterns observed in the baseline Qwen2.5-VL model.

\subsection{Fine-Grained Text-Visual Alignment Metrics}
To quantify visual–language alignment at a finer granularity, we further compare the attention maps $\bar{\mathbf{A}}_i$ for object nouns against the corresponding masks $M_i$ using four common metrics: Average Precision (AP), Area Under Curve (AUC), Normalized Scanpath Saliency (NSS), and Precision. For each metric, we derive binary predictions from $\bar{\mathbf{A}}_i$ using a fixed threshold of $0.75$ if confusion matrix components are required.

\begin{figure}[t] 
  \centering
  \setlength{\abovecaptionskip}{2pt}
  \includegraphics[width=0.95\linewidth]{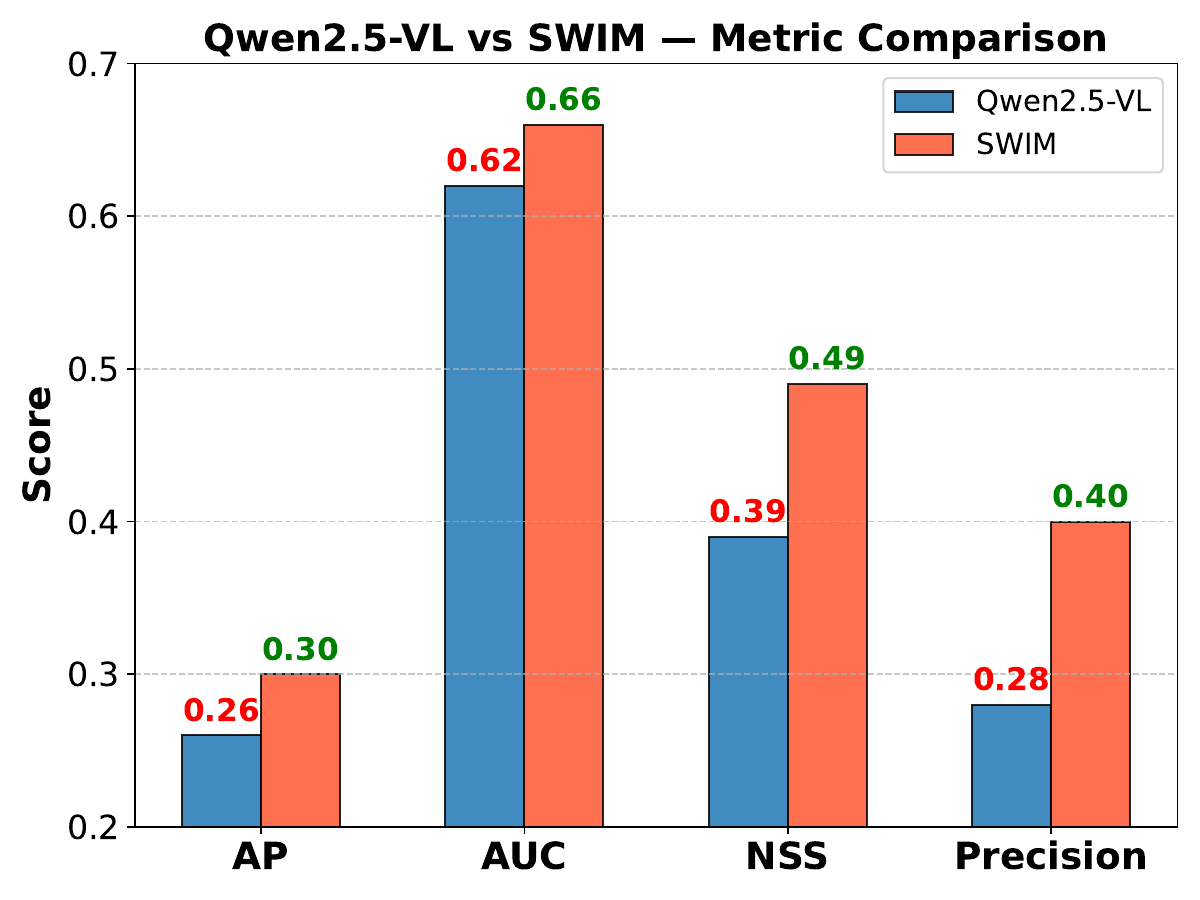}

  \caption{Quantitative comparison of fine-grained text–visual alignment metrics. Evaluation includes AP, AUC, NSS, and Precision. SWIM consistently outperforms the Qwen2.5-VL baseline on all four metrics, indicating more accurate attention on target, fewer false activations, and stronger alignment stability.
  } 
  \vspace{-5pt}
  \label{fig:metric} 
\end{figure}

As shown in \figref{fig:metric}, SWIM consistently outperforms the Qwen2.5-VL baseline across all four metrics (AUC: $0.62 \rightarrow 0.67$, NSS: $0.39 \rightarrow 0.50$, Precision: $0.28 \rightarrow 0.39$, AP: $0.26 \rightarrow 0.30$). These improvements indicate that SWIM generates attention maps with more precise and concentrated coverage of target regions, reduces false activations, and achieves stronger discriminability across thresholds, thereby enhancing fine-grained text–visual grounding.

\subsection{Qualitative Comparisons} 
As demonstrated in \figref{fig:qualitative}, we conduct qualitative comparisons between SWIM and Qwen2.5-VL~\cite{Qwen2.5-VL} on examples that require precisely object reference through natural language. For example, in the caption case, Qwen2.5-VL disregards the prompt's explicit reference and instead describes the most visually salient object in the scene while  SWIM adheres to the prompt and focus on the specified object. 
The other two examples also show cases where SWIM’s outputs match the prompt’s described referent. 
\figref{fig:qualitative} indicates that SWIM’s outputs align more closely with the objects mentioned in the prompts.

\section{Conclusions}
In this paper, we propose SWIM, a training paradigm that applies explicit supervision to improve cross-modal alignment between object nouns and visual regions, thereby enhancing fine-grained object understanding in MLLMs. To enable such supervision, we construct NL-Refer, a refined video dataset with natural language object references paired with mask annotations. SWIM requires no architectural changes and does not need any visual prompts during inference. Experiments show that it achieves state-of-the-art results on fine-grained understanding benchmarks while maintaining competitive performance on general benchmarks. Extensive quantitative analysis further verifies that SWIM can achieve better fine-grained video understanding.

\section{Acknowledgments} This work was supported by NSFC (62522607, 62495061, and 62276145), and the Fundamental Research Funds for the Central
Universities (Nankai University).

{
    \small
    \bibliographystyle{ieeenat_fullname}
    \bibliography{main}
}

\newpage
\appendix

\twocolumn[{%
    \renewcommand\twocolumn[1][]{#1}%
    \begin{center}
        \vspace{-3mm}
        \captionof{table}{GamePoint@K comparison.} 
        \vspace{-3mm}
        \setlength{\tabcolsep}{3pt} 
        \renewcommand{\arraystretch}{1.0}
        \begin{tabular}{lccccc} 
            \toprule
            \textbf{Method} & \textbf{GamePoint@K-1} &  \textbf{GamePoint@K-5} & \textbf{GamePoint@K-10}  & \textbf{GamePoint@K-50} & \textbf{GamePoint@K-100} \\
            \midrule
            Qwen2.5-VL-7B & 0.330 & 0.328 & 0.331 & 0.330 & 0.329 \\
            \textbf{SWIM} & \textbf{0.373} & \textbf{0.375} & \textbf{0.374} & \textbf{0.373} & \textbf{0.374}\\
            \bottomrule
        \end{tabular}

        \label{tab:gamepointK}
    \end{center}%
}]

\section*{Appendix}

\section{Benchmarks}
For completeness, we provide detailed descriptions of the general benchmarks used in the main paper.

\textbf{ActivityNet-QA}~\cite{caba2015activitynet} is a large-scale video question answering benchmark constructed from the ActivityNet dataset. It contains human-annotated question–answer pairs focusing on action-related content, with an average video duration of about 2 minutes. The questions are designed to require understanding of dynamic scenes and temporal sequences rather than static visual cues.

\textbf{VideoMME}~\cite{fu2024videomme} collects videos from a wide range of domains, including sports, documentaries, instructional content, and entertainment. Video durations vary from minutes to hours, making it one of the most comprehensive and challenging benchmarks for holistic video understanding. The diversity in topic, style, and duration tests a model’s ability to handle long-context reasoning and adapt to varied visual–text scenarios.

\textbf{MVBench}~\cite{li2024mvbench} is a multi-choice video understanding benchmark comprising 20 distinct tasks. Each task presents a multiple-choice question targeting temporal comprehension, covering scenarios such as event ordering, cause–effect reasoning, motion tracking, and activity prediction. These tasks require sophisticated temporal reasoning and understanding of dynamic content that cannot be solved by analyzing a single frame, thereby evaluating a model’s capability to integrate information across time.

Together, these benchmarks provide a diverse evaluation landscape: ActivityNet-QA and VideoMME emphasize broad video understanding with varying domain coverage and length, whereas MVBench focuses on fine-grained temporal reasoning across multiple types of challenges.
\section{More Experimental Analysis}

\subsection{GamePoint@K}
We further evaluate retrieval accuracy using GamePoint@K, which measures the fraction of relevant elements among the top-$K$ highest-scoring positions in the attention map:
\begin{equation}
\mathrm{GamePoint@K} = \frac{1}{N} \sum_{i=1}^{N} \frac{|\mathrm{TopK}(\bar{\mathbf{A}}_i, K) \cap P_i|}{|\mathrm{TopK}(\bar{\mathbf{A}}_i, K)|},
\end{equation}
where $\mathrm{TopK}(\bar{\mathbf{A}}_i, K)$ selects the $K$ highest-scoring elements for sample $i$, and $P_i$ denotes its ground-truth positions. Higher GamePoint@K scores indicate that relevant visual targets are ranked closer to the top, reflecting better alignment between textual references and visual regions.

As shown in Table~\ref{tab:gamepointK}, SWIM consistently outperforms Qwen2.5-VL across all $K$ values. At $K=1$, SWIM achieves \textbf{0.373} compared to 0.330 for Qwen2.5-VL, indicating stronger ability to position the correct target at the top rank. This advantage is maintained at broader retrieval depths, with SWIM reaching \textbf{0.375} at $K=5$ (+4.7\% over baseline) and retaining stable gains for $K=10$, $K=50$, and $K=100$. The consistent margins across different $K$ suggest that SWIM produces reliable ranking distributions, keeping relevant objects prominent even as the retrieval list expands.

\subsection{Robustness to Synonym-based Linguistic Noise}
To assess the robustness of SWIM to variations in referring expressions, we conduct an evaluation in which words enclosed in \texttt{<ins>} tags within the VideoRefer-Bench-D prompts are replaced by semantically equivalent synonyms. This modification leaves the overall meaning unchanged but alters the surface form of the text, introducing lexical noise that may challenge models relying on exact token matches. Such a setting reflects real-world scenarios where object references may vary due to differences in speaker style, domain-specific terminology, or translation artifacts, and tests whether a model can preserve grounding accuracy under these conditions. As shown in Table~\ref{tab:synonym}, the original SWIM achieves an average score of 3.78, while SWIM$^*$ obtains 3.74 under synonym perturbations—a marginal drop of 0.04. Compared to Qwen2.5-VL, SWIM maintains strong performance under synonym substitutions, achieving an average accuracy of \textbf{3.74} against 3.43.  These results indicate that SWIM’s alignment mechanism is resilient to changes in word choice, preserving its ability to ground natural language object references to the correct visual regions even under lexical variation.

\begin{table}[ht]
\centering
\setlength{\abovecaptionskip}{2pt}
\small
\tablestyle{6.3pt}{1.0}
\caption{Performance comparisons on VideoRefer-Bench-D. $^*$ denotes incorporating synonym-based noise.}
\begin{tabular}{lcccccccccccc}
\toprule
 &   & \multicolumn{5}{c}{\textbf{VideoRefer-Bench-D}} \\
\cmidrule(lr){2-6} 
\textbf{Method}
 & \textbf{SC} & \textbf{AD} & \textbf{TD} & \textbf{HD} & \textbf{Avg.} \\
\midrule
Qwen2.5-VL-7B~\cite{Qwen2.5-VL} & 3.99 & 3.05 & 2.44 & 2.44 & 2.97 \\
Qwen2.5-VL-7B$^*$~\cite{Qwen2.5-VL} & 4.78 & 3.49	& 3.27 & 2.18 & 3.43\\

  \textbf{SWIM}  & \textbf{4.92} & \textbf{3.85} & \textbf{3.43} & 2.96 & \textbf{3.78}\\
 \textbf{SWIM$^*$} & \textbf{4.86} & \textbf{3.78} & \textbf{3.36} & 2.96 & \textbf{3.74}\\
\bottomrule
\end{tabular}
\label{tab:synonym}
\end{table}
\end{document}